\pdfoutput=1

\documentclass[10pt,twocolumn,letterpaper]{article}

\usepackage[pagenumbers]{iccv} 

\usepackage{amsmath,amsfonts}
\usepackage{algorithmic}
\usepackage{algorithm}
\usepackage{array}
\usepackage{textcomp}
\usepackage{stfloats}
\usepackage{url}
\usepackage{verbatim}
\usepackage{graphicx}

\usepackage{times}
\usepackage{epsfig}
\usepackage{float}
\usepackage{wrapfig}
\usepackage{amssymb,amsthm}
\usepackage{bm,xspace}
\usepackage{comment}
\usepackage{verbatim}
\usepackage{multirow}
\usepackage{balance}
\usepackage{booktabs}
\usepackage{etoolbox,siunitx}
\usepackage{calc}
\usepackage{pifont,hologo}
\usepackage{nicefrac}
\usepackage{siunitx}
\usepackage{colortbl}

\usepackage[dvipsnames]{xcolor}

\def\secref#1{Sec.~\ref{#1}}
\def\figref#1{Fig.~\ref{#1}}
\def\tabref#1{Tab.~\ref{#1}}
\def\eqref#1{Eq.~\ref{#1}}
\def\algref#1{Alg.~\ref{#1}}

\makeatletter
\DeclareRobustCommand\onedot{\futurelet\@let@token\@onedot}
\def\@onedot{\ifx\@let@token.\else.\null\fi\xspace}
\def\eg{\emph{e.g}\onedot} 
\def\ie{\emph{i.e}\onedot}

\def\etal{\emph{et~al}\onedot}
\def\etalcite#1{\etal~\cite{#1}}
\makeatother

\DeclareSIUnit\pixel{px}

\definecolor{iccvblue}{rgb}{0.21,0.49,0.74}
\usepackage[pagebackref,breaklinks,colorlinks,allcolors=iccvblue]{hyperref}

\def\paperID{5569} 
\def\confName{ICCV}
\def\confYear{2025}

\title{A Linear N-Point Solver for Structure and Motion from Asynchronous Tracks}

\author{Hang Su$^{1}$\thanks{indicates equal contribution} \quad Yunlong Feng$^{1}$\footnotemark[1] \quad Daniel Gehrig$^{2}$ \quad Panfeng Jiang$^{1}$ \quad Ling Gao$^{3}$ \\ \quad Xavier Lagorce$^{1}$ \quad Laurent Kneip$^{1,4}$\\ \\
$^1$ ShanghaiTech University,
$^2$ University of Pennsylvania,
$^3$ Amap, Alibaba Group, \\
$^4$ Shanghai Engineering Research Center of Intelligent Vision and Imaging
}

\begin{document}
\maketitle

\begin{abstract}
Structure and continuous motion estimation from point correspondences is a fundamental problem in computer vision that has been powered by well-known algorithms such as the familiar 5-point or 8-point algorithm. However, despite their acclaim, these algorithms are limited to processing point correspondences originating from a pair of views each one representing an instantaneous capture of the scene. 
Yet, in the case of rolling shutter cameras, or more recently, event cameras, this synchronization breaks down. In this work, we present a unified approach for structure and linear motion estimation from 2D point correspondences with arbitrary timestamps, from an arbitrary set of views. By formulating the problem in terms of first-order dynamics and leveraging a constant velocity motion model, we derive a novel, linear point incidence relation allowing for the efficient recovery of both linear velocity and 3D points with predictable degeneracies and solution multiplicities. Owing to its general formulation, it can handle correspondences from a wide range of sensing modalities such as global shutter, rolling shutter, and event cameras, and can even combine correspondences from different collocated sensors. We validate the effectiveness of our solver on both simulated and real-world data, where we show consistent improvement across all modalities when compared to recent approaches. We believe our work opens the door to efficient structure and motion estimation from asynchronous data. Code can be found at \url{https://github.com/suhang99/AsyncTrack-Motion-Solver}.
\end{abstract}

\section{Introduction}
\label{sec:intro}

\begin{figure}
    \centering
    \includegraphics[width=0.75\linewidth]{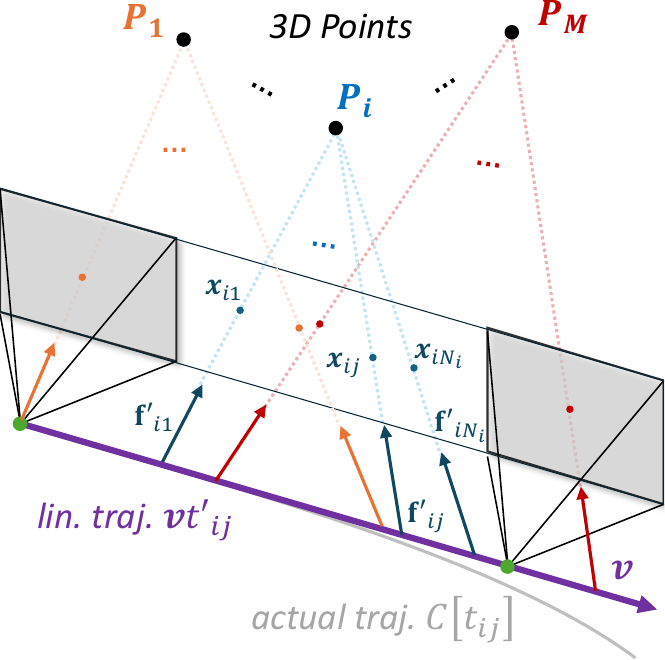}
    \vspace{-2ex}
    \caption{We develop a linear N-point solver for recovering 3D points $\mathbf{P}_i$ and the velocity $\mathbf{v}$ of a camera undergoing quasi-linear motion, given a set of timestamped observations $(\mathbf{x}_{ij},t_{ij})$. No assumptions are made about the temporal synchronization of observations $\mathbf{x}_{ij}$, yielding a general algorithm that can handle observations from global shutter, rolling-shutter, and fully asynchronous rolling sensors, such as event cameras. Observations $\mathbf{x}_{ij}$ are converted into rotation compensated bearing vectors $\mathbf{f}'_{ij}\doteq \text{exp}\left(\left[\omega t'_{ij}\right]_\times\right)\mathbf{K}^{-1}\mathbf{\tilde x}_{ij}$, and used to construct a set of linear point incidence relations. Here $t'_{ij}=t_{ij}-t_s$ denotes relative time, $t_s$ the reference time, $\mathbf{K}$ the camera intrinsics, $\boldsymbol{\omega}$ the angular rate, which can be given by an IMU or upstream estimation algorithm. }
    \label{fig:asynchronous-features}
    \vspace{-3ex}
\end{figure}

Finding the continuous motion of a single monocular camera is one of the most fundamental problems in the area of geometric computer vision. In the calibrated case, the core of the solution to the visual odometry problem entails the identification of the \emph{extrinsic parameters} relating a pair of views of a scene, meaning a Euclidean transformation consisting of a relative orientation and an up-to-scale translational displacement. Note however that in the video-based, continuous motion setting, there is only a marginal difference between finding the relative pose and finding local camera dynamics. Indeed, if the images come from a constant-rate video stream, we merely have to divide the relative displacement variables by the frame sampling period to obtain approximate local camera dynamics.\footnote{The matter may complicate if more jerky stop-and-go motion is involved, in which case key-frame selection may be required, and the optimization problem typically transitions into a multi-view, absolute pose bundle adjustment scheme. However, the statement is approximately valid under the assumption of a simple, continuous trajectory.}

It might not seem obvious, but the primary reason why---in the classical visual odometry scenario---we stick to relative transformation parameters rather than first-order dynamics is the very nature of the sampling mechanism of traditional cameras: Images are sampled synchronously, and---at least in the common case---at a relatively low rate, which easily tends to temporally under-sample more agile camera motion. For this reason, motion estimation is traditionally framed as the recovery of delta transformations, instead of local first-order dynamics. However, with the introduction of temporally denser and notably asynchronously sampling sensors , the consideration of first-order dynamics and a constant velocity motion model becomes a practical necessity.

Important examples of such sensors are given by high-speed cameras, rolling shutter cameras, and event cameras. Rolling shutter cameras notably capture images one row at a time, in a quasi-asynchronous way, leading to timestamp differences at different rows of the image, and these differences need to be meaningfully addressed in a given motion estimation algorithm. The event camera on the other hand is a relatively new type of visual sensor~\cite{lichtensteiner2008128} and has been a recent enabler of high-speed and low-power computer vision due to its unique working principle. It consists of independent pixels that return an asynchronous stream of per-pixel brightness change measurements, \textit{i.e. events}. Each event indicates a discrete, signed change in perceived brightness timestamped by a clock with microsecond-level resolution. 

In this work, we focus on motion estimation using geometric feature-based methods. While a plethora of such methods exist for global shutter and rolling shutter cameras, none manage to natively support features extracted at asynchronous timestamps from potentially row-wise or completely asynchronous sensors. 
We fill this gap by building upon a recently introduced geometric method for line feature-based motion and structure estimation from asynchronous measurements~\cite{gao2024eventail,gao2023eventail}, and extend it to operate on 3D points instead. Originally proposed for event cameras, the solver utilizes a small, sparse set of events generated anywhere along the reprojection of a straight line perceived under constant linear motion. Surprisingly, given sufficient measurements of a single line, it is possible to recover the full 3D location of the straight line as well as a projection of the 3D camera velocity.

The present work makes the following contributions:
\begin{itemize}
    \item We propose a new solver for geometric motion and structure estimation from asynchronous, time-stamped reprojections of points measured under the assumption of constant linear motion with known rotation. We thus extend previous line-feature-based solvers by a novel, point-feature based approach, and contribute to a growing theoretical body of geometric incidence relations that operate over dynamics and measurements sampled arbitrarily in space and time.
    \item The proposed method is a single-stage linear solver that operates over an arbitrary number of asynchronous point feature tracks and is highly efficient by employing the Schur complement trick. We furthermore outline the exact conditions under which the full linear velocity becomes observable. Surprisingly, under a linear motion model, three temporal observations of only a single point enable us to recover the full orientation of the displacement baseline as well as the corresponding 3D point.
    \item Through our experiments, we demonstrate general applicability of the theory to various types of cameras, including regular global shutter cameras, rolling shutter cameras, and event cameras.
\end{itemize}

\section{Related Work}
\label{sec:related_work}

\noindent\textbf{Geometric Solvers:} The geometric solution of camera motion is undoubtedly one of the major success stories of computer vision. Based on epipolar geometry~\cite{hartley2003multiple}, a plethora of algorithms has been proposed to efficiently estimate scene properties and camera parameters from a set of 2D point correspondences (\ie~pixel observations) between two views~\cite{Nister2004fivepoint,stewenius06}. 

Relaxing the synchronicity assumption is already required when addressing correspondences captured from rolling shutter images. Important theory has been devised to estimate absolute pose and motion from known 2D-3D correspondences using either linear or polynomial solvers~\cite{Saurer15iros,Kukelova18accv,Albl20tpami,kukelova2020minimal}. In particular, Saurer~\etalcite{Saurer15iros} use a similar point incidence relation as the one presented in this work. Unlike these works, our method uses only 2D point measurements and simultaneously solves motion and structure via a linear system.
The work in~\cite{Lao21pami} targets homography estimation from point correspondences derived from rolling shutter images, but requires the observed 3D points to be on a plane. 
Dai~\etalcite{Dai16cvpr} also relax the synchronicity assumption, however using a pair-wise epipolar co-planarity constraint instead of the N-point incidence relation proposed in this work. Further related approaches are given by n-linearities~\cite{hartley2003multiple}, visual-inertial bootstrapping approaches~\cite{kneip2011closed}, or recent works on visual odometry on cars~\cite{huang2019motion}. However---though able to process N-point feature tracks---these approaches are proposed for regularly sampling global shutter cameras, or simply multiple individual cameras. To the best of our knowledge, we propose the first theory that relies on a constant velocity motion model and permits for fully asynchronous measurements.\\

\noindent\textbf{Event-based Motion Estimation:} Various methods for feature extraction and tracking---both data-driven~\cite{hamann2025motion,messikommer23cvpr,Messikommer25tpami}, and traditional~\cite{lagorce2015spatiotemporal,gehrig2020eklt,gehrig2018asynchronous}---as well as visual odometry~\cite{mueggler2018continuous,liu2021spatiotemporal,vidal2018ultimate,Pellerito_2024_IROS,klenk2023devo,kim2016real,rebecq2016evo,zhu2017event} have already been proposed. A common strategy in many of these works, particularly in learning-based pipelines, is to aggregate events into synchronous, frame-like representations. This approach, however, sidesteps the advantage of event cameras. In parallel, some earlier research explicitly exploit the asynchronous nature of the data, developing methods for motion estimation that operate directly on the event stream. Despite these different approaches, the critical question of closed-form bootstrapping often remains unaddressed.
An interesting alternative for event-based motion and structure estimation that processes raw events with their original timestamp is given by contrast maximization~\cite{gallego2018unifying,stoffregen2019event1,peng2022globally}. By employing a compact parametric image warping function, events are unwarped into a reference view in which---owing to the sparsity of strong appearance gradients---the entropy of their distribution is minimized. Although the framework has been successfully applied to various problems such as motion estimation, optical flow, and depth estimation~\cite{hamann2025motion,kim2021real,liu2020globally,shiba2024secrets}, it is a computationally intensive approach that involves iterative batch optimization over many events and remains restricted to homographic warping scenarios (\eg~pure rotation, planar scene).

Several works~\cite{lu2023event,niu2024esvo2,chamorro2023event} also combine event cameras with IMU sensors to address high-speed maneuvers and challenging conditions, leveraging the complementary properties of both sensing modalities.
Recent approaches have explored efficient, sparse geometric solvers better suited to the asynchronous nature of event data. Peng~\etalcite{peng2021continuous} and Xu~\etalcite{xu2024tight} utilize three-view geometry based on 3D lines for camera ego-motion estimation. Gao~\etalcite{gao2023eventail,gao2024eventail} improve this idea by developing an N-point linear solver for line and motion estimation, providing new insights into the manifold distribution of the events generated by the observation of a straight line under motion. Built upon this, Zhao~\etalcite{zhao2025full} propose a new solver for full-DoF motion estimation via rank minimization. Unlike these approaches, our work focuses on asynchronous point feature tracks and exploits their spatio-temporal characteristics for accurate motion estimation.
\section{Methodology}
\label{sec:method}

We model the observations of $M$ 3D points $\{\mathbf{P}_i\}_{i=1}^M$ by a calibrated camera undergoing an arbitrary 6 degrees of freedom (DoF) motion on the time interval $\mathcal{T}=[t_s-\Delta t,t_s + \Delta t]$, with reference time $t_s$, and half-width $\Delta t$. We denote the camera pose at time $t\in\mathcal{T}$ with $\mathbf{C}(t)\in SE(3)$. Throughout this duration we assume each point $\mathbf{P}_i\in\mathbb{R}^3$ to be observed $N_i$ times by a point tracker, leading to spatio-temporal track observations $\mathcal{X}_i=\{(\mathbf{x}_{ij}, t_{ij})\}_{j=1}^{N_i}$ in the image plane. Each observation $(\mathbf{x}_{ij}, t_{ij})$ comprises the projection $\mathbf{x}_{ij}$ of point $\mathbf{P_i}$ at timestamp $t_{ij}$ in the image plane. 

Note that no assumption is made on the synchronicity of timestamps $t_{ij}$, leading to a very general formulation. This formulation can handle a wide range of tracking scenarios and sensing modalities: 
\begin{enumerate}
    \item Tracks derived from a sequence of global shutter images, which may or may not be temporally aligned due to loss of tracking or track initialization.
    \item Row-wise synchronized tracks derived from a rolling shutter camera, where points at different row coordinates do not necessarily share the same timestamp $t_{ij}$.
    \item Fully asynchronous feature tracks from an event camera. These tracks may be densely sampled in time and have little to no timestamp coherence. 
\end{enumerate}
We will show that our solver seamlessly handles all of these cases. In~\secref{sec:incidence}, we will present the relevant incidence relation, which forms a set of linear constraints between the point location, camera motion, and point observation. Then, in~\secref{sec:linear} we will present how to write a set of such constraints as a linear system, before solving it in~\secref{sec:solver}. We discuss properties of the solver in~\secref{sec:properties}. \secref{sec:implementation} concludes with implementation details.

\subsection{Incidence Relationship}
\label{sec:incidence}

We will assume a configuration as illustrated in~\figref{fig:asynchronous-features}, where a camera with pose $\mathbf{C}(t)$ composed of orientation $\mathbf{R}(t)$ and position $\mathbf{p}(t)$ undergoes quasi-linear dynamics on the small interval $\mathcal{T}=[t_s-\Delta t, t_s+\Delta t$]. 

Let $\mathbf{\tilde x}_{ij}$ be the 3D homogeneous coordinate of $\mathbf{x}$ and $\mathbf{f}_{ij}=\mathbf{K}^{-1}\mathbf{\tilde x}_{ij}$ be the normalized coordinates (\ie~bearing vectors) of the feature tracks previously described. 
Our incidence relation leverages the fact that the 3D point $\mathbf{P}_i$ observed at time $t_{ij}$ should project onto the observed point $\mathbf{f}_{ij}$. We find the 3D point in camera coordinates $\mathbf{P}'_{ij}$ at time $t_{ij}$ as 
\begin{equation}
    \mathbf{P}'_{ij} = \mathbf{R}^\intercal(t_{ij})\left(\mathbf{P}_{i} - \mathbf{p}(t_{ij})\right) \,.
\end{equation}
The above constraint implies that $\mathbf{f}_{ij}$ and $\mathbf{P}'_{ij}$ are parallel, which can be formulated as a constraint on their cross-product 
\begin{equation}
    \mathbf{f}_{ij}\times \left(\mathbf{R}^\intercal(t_{ij})\left(\mathbf{P}_{i} - \mathbf{p}(t_{ij})\right)\right) = \mathbf{0}_{3\times 1} \,.
\end{equation}
Using properties of cross products and introducing the rotated bearing $\mathbf{f}'_{ij}\doteq\mathbf{R}(t_{ij})\mathbf{f}_{ij}$ yields
\begin{equation}
    \mathbf{f}'_{ij}\times \left(\mathbf{P}_{i} - \mathbf{p}(t_{ij})\right) = \mathbf{0}_{3\times 1} \,,
\end{equation}
which is our desired incidence relation, also treated in \cite{Saurer15iros}. In the appendix we show that this incidence relation can be specialized to the epipolar constraint used in the familiar 5-point or 8-point algorithm~\cite{hartley2003multiple} or the line incidence relation used in the recent line solver in~\cite{gao2023eventail,gao2024eventail}. In the next section, we will describe how to use this relation to solve for the 3D points and camera motion. 

\subsection{Transition to Linear System}
\label{sec:linear}

Expressing all quantities with respect to the reference frame $\mathbf{C}(t_s)=\mathbf{I}_{4\times 4}$ at time $t_s$ we expand the motion of the camera with a Taylor Series $\mathbf{R}(t_{ij})\approx \text{exp}(\left[\boldsymbol{\omega}t'_{ij}\right]_\times)$ and $\mathbf{p}(t_{ij})\approx \mathbf{v}t'_{ij}$. We introduce angular rate $\boldsymbol{\omega}$, linear velocity $\mathbf{v}$ and relative timestamp $t'_{ij} = t_{ij} - t_s$. The operation $[.]_\times$ maps vectors to a skew-symmetric matrix. 
Note that the degree of expansion is arbitrary, and each chosen degree will yield a given system of equations that are linear in the 3D points and body rates. In what follows, however, we will focus on a linear expansion, and will present the arbitrary case in the appendix, and applications in the experiments.

As done in previous work, we focus on finding the linear velocity $\mathbf{v}$ for a given $\boldsymbol{\omega}$, which we assume to be given either by an external IMU (as in~\cite{gao2024eventail}), or other rotation estimation algorithms~\cite{gallego2017accurate,zhao2025full}. 

We make use of the fact that the cross product with $\mathbf{f}'_{ij}$ can be rewritten as a product with $\left[\mathbf{f}'_{ij}\right]_\times$ 
\begin{equation}
    \left[\mathbf{f}'_{ij}\right]_\times \mathbf{P}_{i} - t'_{ij}\left[\mathbf{f}'_{ij}\right]_\times\mathbf{v} = \mathbf{0}_{3\times 1} \,.
\end{equation}
We gather all such constraints that involve the point $\mathbf{P}_i$ into a single system of equations.
\begin{equation}
\label{eq:one_track}
    \underbrace{\begin{bmatrix}
        \left[\mathbf{f}'_{i1}\right]_\times & -t'_{i1} \left[\mathbf{f}'_{i1}\right]_\times \\ 
        \vdots & \vdots \\
        \left[\mathbf{f}'_{iN_i}\right]_\times & -t'_{iN_i} \left[\mathbf{f}'_{iN_i}\right]_\times
    \end{bmatrix}}_{\doteq\begin{bmatrix}
        \mathbf{F}_i&\mathbf{G}_i
    \end{bmatrix}\in\mathbb{R}^{3N_i\times 6}} \begin{bmatrix}
        \mathbf{P}_i\\\mathbf{v}
    \end{bmatrix}=\mathbf{0}_{3N_i\times 1} \,.
\end{equation}
In a last step, we stack all such constraints originating from different points $\mathbf{P}_i$ into one large system yielding 
\begin{equation}
    \label{eq:multi-point-incidence}
    \underbrace{
    \begin{bmatrix}
        \mathbf{F}_1 & & & & \mathbf{G}_1 \\
        & \mathbf{F}_2 & & & \mathbf{G}_2 \\
        & & \ddots & & \vdots \\
        & & & \mathbf{F}_M & \mathbf{G}_M
    \end{bmatrix}}_{\doteq \mathbf{A}\in \mathbb{R}^{3N\times (3M+3) }}
    \underbrace{\begin{bmatrix}
        \mathbf{P}_1 \\
        \mathbf{P}_2 \\
        \vdots \\
        \mathbf{P}_M \\
        \mathbf{v}
    \end{bmatrix}}_{\doteq \mathbf{x}\in\mathbb{R}^{3M+3}}
    = \mathbf{0}_{3N\times1} \,,
\end{equation}
where we call $N=\sum_i N_i$ the total number of observations. 
Finally, we notice that this system imposes a linear constraint on the unknown points $\mathbf{P}_i$ and camera velocity $\mathbf{v}$, and thus it admits an efficient solver, discussed next. 
\subsection{Solver}
\label{sec:solver}
The linear system above could be solved with standard tools, by employing a singular value decomposition (SVD) on the matrix $\mathbf{A}$ and then recovering the last column of the orthogonal matrix $\mathbf{V}$ corresponding to the smallest singular value of $\mathbf{A}$. However, as the number of observations increases, computing the singular value decomposition may start to pose a computational burden. For this reason, we first limit our focus to only recovering $\mathbf{v}$, and then show how to find the $\mathbf{P}_i$. As we will see, the sparse structure of $\mathbf{A}$ allows the derivation of an efficient solver. We start off by left multiplying the linear system by $\mathbf{A}^\intercal$, and writing the resulting system as a block system of equations:
\begin{equation}
    \label{eq:schur-complement}
    \underbrace{\mathbf{A}^\intercal\mathbf{A}}_{\doteq \mathbf{M}}\mathbf{x}=
    \begin{bmatrix}
        \mathbf{M}_{A} & \mathbf{M}_{B} \\
        \mathbf{M}_{B}^\intercal & \mathbf{M}_{D}
    \end{bmatrix}
    \begin{bmatrix}
        \mathbf{P}_{1:M} \\
        \mathbf{v}  
    \end{bmatrix} 
    = \mathbf{0}_{(3M+3)\times 1} \,,
\end{equation}
where we have stacked $\mathbf{P}_i$ into $\mathbf{P}_{1:M}$, and the dimensions of the subblocks are $\mathbf{M}_A\in\mathbb{R}^{3M\times 3M}$, $\mathbf{M}_B\in\mathbb{R}^{3M\times 3}$ and $\mathbf{M}_D\in\mathbb{R}^{3\times 3}$. We write out the explicit form of these matrices in the appendix. We then employ the Shur-complement trick to write a system only in $\mathbf{v}$ which has the form 
\begin{equation}
    \label{eq:solve-v}
    \underbrace{(\mathbf{M}_{D} - \mathbf{M}_{B}^\intercal\mathbf{M}_{A}^{-1}\mathbf{M}_B)}_{\doteq \mathbf{B}\in\mathbb{R}^{3\times 3}}\mathbf{v} = \mathbf{0}.
\end{equation}
This last equation can be efficiently solved by employing SVD on the matrix $\mathbf{B}$, finding the normalized velocity estimate $\hat{\mathbf{v}}$ as the principle direction corresponding with the smallest singular value of $\mathbf{B}$. Note that the velocity is normalized due to absence of scale in a monocular setup.

One may think that the inversion of $\mathbf{M}_A$ in~\eqref{eq:solve-v} is expensive since $\mathbf{M}_A\in\mathbb{R}^{3M\times 3M}$ leading naively to $O(M^3)$ complexity. However, the matrix $\mathbf{M}_A$ is actually block diagonal with $M$ blocks of size $3\times 3$, leading to an efficient inversion algorithm of complexity $O(M)$ instead. Moreover, all terms can be computed from a linear combination of terms $[\mathbf{f}'_{ij}]^2_\times =\mathbf{f}'_{ij}{\mathbf{f}'_{ij}}^\intercal-\Vert \mathbf{f}'_{ij}\Vert^2 \mathbf{I}_{3\times 3} $, leading to significant sharing of computation.
Finally, having found the estimate $\hat{\mathbf{v}}$ we can find the solution to $\mathbf{P}_i$ as 
\begin{equation}
    \label{eq:solve-p}
    \hat{\mathbf{P}}_i = -(\mathbf{F}_i^\intercal\mathbf{F}_i)^{-1}\mathbf{F}_i^\intercal\mathbf{G}_i\hat{\mathbf{v}} \,,
\end{equation}
which can be done efficiently by reusing computation from~\eqref{eq:solve-v}. Now let us analyze the properties of our solver, how many solutions it generates, and when it may fail. 

\subsection{Solver Properties}
\label{sec:properties}

\textbf{Solution Multiplicity:} We start off by discussing the solution multiplicity of the above solver. The SVD operation in~\eqref{eq:solve-v} yields two possible unit vectors $\hat{\mathbf{v}}$ and $-\hat{\mathbf{v}}$ which entail the two possible solutions $\hat{\mathbf{P}}_{1:M}$ and $-\hat{\mathbf{P}}_{1:M}$. During deployment, we select the correct solution by recognizing that the recovered points $\hat{\mathbf{P}}_i$ must have positive depth. We thus test for the following condition
\begin{equation}
\label{eq:check}
    (\hat{\mathbf{P}}_i)_z = -((\mathbf{F}_i^\intercal\mathbf{F}_i)^{-1}\mathbf{F}_i^\intercal\mathbf{G}_i\mathbf{\hat{v}})_z > 0
\end{equation}
and invert the velocity if it is violated.

\noindent\textbf{Degeneracy:} Next, let us analyze when degenerate solutions may be encountered. We see that the solution of $\mathbf{v}$ in~\eqref{eq:solve-v} depends on the inversion of $\mathbf{M}_A$ which in turn depends on the inversion of block diagonal matrices of the form $\mathbf{F}_i^\intercal \mathbf{F}_i$. To successfully invert these matrices, we require that $\mathbf{F}_i$ has a full rank. $\mathbf{F}_i$ is composed of matrices $[\mathbf{f}'_{ij}]_\times $ which only have two independent rows. We may thus consider the reduced form of $\mathbf{F}_i$ with size $2N_i\times 3$ where we have deleted every third row. Thus, to have full rank $N_i\geq 2$, \ie~we must enforce that every track has at least two different observations. In practice, we do this by simply discarding tracks with only one observation. Finally, enforcing  $\text{rank}(\mathbf{B})\geq 2$ ensures that the SVD step succeeds in~\eqref{eq:solve-v}. For a discussion on the rank of $\mathbf{B}$ see the appendix.

\noindent\textbf{Constraint Analysis:} The system in~\eqref{eq:multi-point-incidence} has $3M+2$ unknowns (number of variables minus one for unobservability of scale), and 3N constraints. However, each $[\mathbf{f}'_{ij}]_\times$ only adds two linearly independent constrains, so the number of constraints is actually $2N$. Thus recovering all unknowns needs
\begin{equation}
    2N\geq 3M+2 \implies N\geq \left\lceil\frac{3M}{2}\right\rceil+1
\end{equation}
observations. As discussed above, we require at least two observations per 3D point for stable inversion, \ie
\begin{equation}
 N\geq 2M   
\end{equation}
We now consider four cases: 
\begin{itemize}
    \item $M=1$: Here $N=N_1\geq 3$ leads to a overconstrained system of at least 6 equations with 5 unknowns. Dropping one equation makes this case minimal.
    \item $M=2$: Here $N\geq4$. In particular $N_1=N_2=2$ leads to \emph{minimal} solver with $8$ equations in $8$ unknowns. For larger $N$ the system becomes overconstrained again.
    \item $M=3$: Here $N\geq6$ with $N_1=N_2=N_3=2$ yielding a \emph{minimal} system of $12$ equations in $12$ unknowns.
    \item $M>3$: Here $N\geq2M$ leading to an overconstrained system of $2N$ equations in $3M+2$ unknowns.
\end{itemize}
Interestingly, the first three cases all give rise to minimal 3 point, 4 point or 6 point algorithms.
We summarize the complete algorithm in~\algref{alg:solver}.

\begin{algorithm}[t!]
    \caption{N-Point Solver for Structure \& Motion}
    \textbf{Input:} A set of track observations $(\mathbf{x}_{ij}, t_{ij})$, and angular rate $\boldsymbol{\omega}$ from an IMU. Reference time $t_s$.\\
    \textbf{Output:} Estimates of points $\hat{\mathbf{P}}_i$ and linear velocity $\hat{\mathbf{v}}$.
    \begin{itemize}
        \item Compute rotated bearing vectors $\mathbf{f}'_{ij}=\text{exp}\left([\omega t'_{ij}]_\times\right)\mathbf{f}_{ij}$.
        \item Compute $\mathbf{F}_i$ and $\mathbf{G}_i$ in~\eqref{eq:one_track}. Ensure that $\text{rank}(\mathbf{F}_i)=3$, otherwise terminate.
        \item Compute $\mathbf{B}$ in~\eqref{eq:solve-v} and solve for $\hat{\mathbf{v}}$. Terminate if the rank of $\mathbf{B}$ is smaller than 2.
        \item Compute $\hat{\mathbf{P}}_i$ from~\eqref{eq:solve-p}. 
        \item Check the depth via inequality~\eqref{eq:check}. If it is negative, invert the signs of $\hat{\mathbf{P}}_i$ and $\hat{\mathbf{v}}_i$.  
    \end{itemize} 
    \label{alg:solver}
\end{algorithm}

\subsection{Implementation Details}
\label{sec:implementation}
We derive feature tracks from a variety of input modalities using off-the-shelf trackers, and then embed the proposed point solver into a RANSAC loop. This loop removes outliers from poor tracking, and produces a refined estimate based on the found inliers~\cite{fischler1981random}.

\noindent\textbf{Feature Tracking:} We use off-the-shelf trackers designed for \emph{(i)} global shutter cameras, \emph{(ii)} rolling shutter cameras, and \emph{(iii)} event cameras. Each tracker provides observations $\mathbf{x}_{ij}$, which are converted into rotated bearing vectors $\mathbf{f}'_{ij}$, based on IMU angular rate readings $\boldsymbol{\omega}$. For global shutter cameras, the timestamp $t_{ij}$ of the observation is simply the image timestamp. For rolling shutter cameras $t_{ij} - \frac{y_{ij}}{(H-1)T_{\text{rs}}}$ is corrected by the row index $y_{ij}$ scaled by the row scanning time $T_{\text{rs}}$, and height in pixels $H$ of the sensor. For event cameras, timestamps are assigned based on measured events, resulting in asynchronous tracks. Tracks shorter than 2 are pruned to avoid degeneracy in the solver.

\noindent\textbf{RANSAC:} In each iteration of RANSAC we perform the following three steps: First, to balance computational complexity with spatio-temporal distribution, we sample $M$ feature tracks, and then $N_i=n\geq 2$ temporally distributed observations $(\mathbf{f}'_{ij},t'_{ij})$. 
Then, we generate a velocity hypothesis $\hat{\mathbf{v}}$ based on the $N=\sum_i N_i$ observations following the solver in~\secref{sec:solver}, and rejecting the solution yielding negative point depth. Next, inliers are identified via the consistency of $\hat{\mathbf{v}}$ with observations $(\mathbf{f}'_{ij},t_{ij})$. For track $i$, we predict the 3D point $\hat{\mathbf{P}}_i$ from~\eqref{eq:solve-p}, map it into the frame at each time $t_{ij}$, resulting in $\hat{\mathbf{P}}'_{ij}=\hat{\mathbf{P}}_i-\hat{\mathbf{v}} t'_{ij}$ and then project it into the current frame, yielding bearing estimate $\hat{\mathbf{f}}'_{ij}$. We use the average angular residual $\bar{\theta}$ between the observed and estimated bearing vectors along the track as error metric
\begin{equation}
    \label{eq:ransac-residual}
    \bar{\theta_i} = \frac{1}{N_i}\sum_{j=1}^{N_i}\arccos\left(\frac{{\mathbf{f}'}_{ij}^\intercal\hat{\mathbf{f}}'_{ij}}{\|\mathbf{f}'_{ij}\|\|\hat{\mathbf{f}}'_{ij}\|}\right)
\end{equation}
A track is classified as an inlier if $\bar{\theta_i}$ is lower than a certain threshold (\eg~$5^{\circ}$), and the hypothesis with the highest number of inliers is retained throughout iteration.

After termination, the inliers corresponding with the best hypothesis are used to estimate the hypothesis $\hat{\mathbf{v}}$ leading to a refined estimate. The inlier ratio serves as a confidence metric reflecting the solver's robustness to outlier tracks.

\section{Experiments}
\label{sec:experiments}

\begin{figure*}
    \centering
    \includegraphics[width=0.9\linewidth]{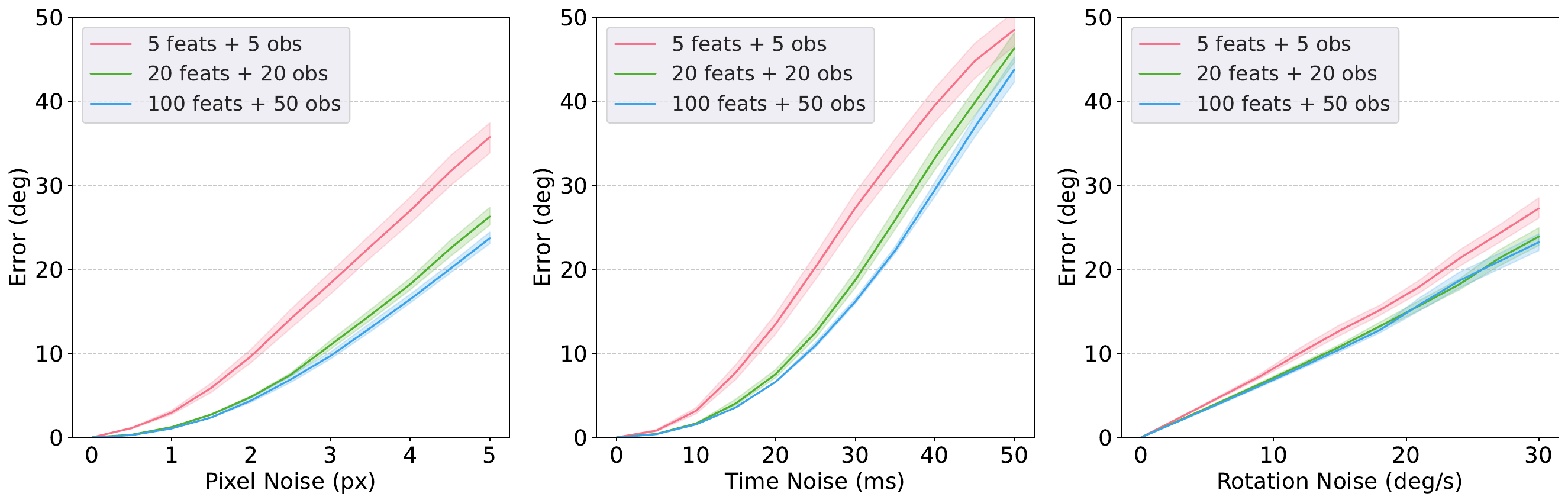}
    \vspace{-3ex}
    \caption{\textbf{Left}: Pixel noise sensitivity; \textbf{Middle}: Timestamp jitter impact; \textbf{Right}: Rotation perturbation effects. Each plot compares three observation levels with the number of features and the number of observation per track: sparse (5-5), moderate (20-20), and dense (100-50). Shaded regions represent error bounds across 1000 trials.}
    \label{fig:sim_noise}
\end{figure*}

We comprehensively evaluate the performance of the proposed point solver, in two stages: First, we validate our method in a simulated environment (\secref{sec:sim-exp}), where we study its sensitivity to different noise source including timestamp jitter, pixel noise and noise on the angular rate readings. We also study its accuracy as a function of the number of tracks and number of observations per track. 

In a second step, we then report results in real world settings (\secref{sec:real-exp}), where we study its application to tracks derived from global-shutter, rolling-shutter and event-based cameras. Throughout the experimental section, we will report the accuracy of the scale-less velocity similar to~\cite{gao2024eventail}, which is defined as the angular error between the true velocity $\mathbf{v}_{\text{gt}}$ and estimate $\hat{\mathbf{v}}$ by
\begin{equation}
    \theta_{\text{err}} = \text{arccos}\left(\frac{\mathbf{v}_{\text{gt}}^\intercal\hat{\mathbf{v}}}{\|\mathbf{v}_{\text{true}}\|\|\hat{\mathbf{v}}\|}\right)
\end{equation}
Errors in 3D point estimation are not reported, as they are typically subsumed in the velocity error. In the appendix, we also apply our method for normalized acceleration estimation, with a similar error metric as above. 

\subsection{Simulation Experiments}
\label{sec:sim-exp}
We set a virtual camera with a resolution of $640\times480$ and a focal length of $320$ pixels. A velocity vector with fixed magnitude $\|\mathbf{v}\| = 1 m/s$ and random direction is generated to simulate camera motion. Static 3D points are randomly distributed within a one-meter cubic volume positioned two meters in front of the camera. This ensures that no points cross the camera plane during the motion. Observations are generated over a sliding time window of 0.2 seconds, with timestamps uniformly sampled within this interval. Each scenario is repeated 1,000 times to ensure statistical significance. Our solver achieves a minimal-case runtime of \SI{63}{\micro\second} on CPU (Intel Xeon Platinum 8352V@3.5GHz). In the following sections, we analyze three key factors: noise resilience, observation count and track length.

\subsubsection{Analysis of Noise Resilience}

We first evaluate the solver's robustness under three noise sources: inaccurate point tracking, temporal misalignment, and orientation drift in camera pose. As illustrated in~\figref{fig:sim_noise}, experiments vary noise level across practical ranges: pixel noise (0 - 5 pixels), timestamp jitter (0 - 50ms), and rotational perturbation (angular velocity noise of 0 - 30 deg/s), while testing three different settings of observation. The results demonstrate that the solver achieves sub-$5^\circ$ error at moderate noise levels (1 pixels, 10 ms jitter, 5 deg/s), validating its feasibility in typical operational scenarios. Notably, timestamp jitter exhibits a nonlinear error escalation, with error rising sharply beyond 15 ms. In contrast, pixel noise induces near-linear error scaling, suggesting the tolerance to common feature-tracking inaccuracies, while rotation perturbation has linear impact on the performance. We also notice more observations can effectively mitigate errors due to pixel noise and timestamp jitter, yet yield limited improvement for rotation-induced errors. Nevertheless, external sensor such as IMU can address this limitation by providing rotation-compensated inputs.

\subsubsection{Analysis of Spatial-temporal Observations}
\begin{figure*}
    \centering
    \includegraphics[width=0.85\linewidth]{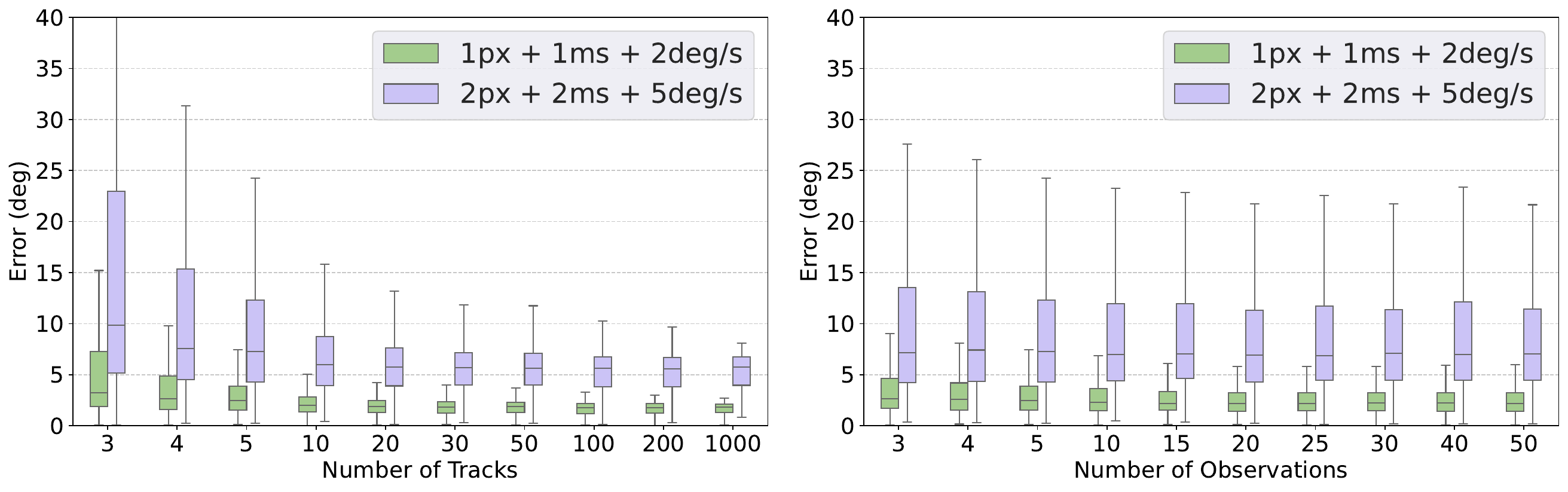}
    \vspace{-3ex}
    \caption{Track count (\textbf{left}) vs. Observation count per track (\textbf{right}). Box colors indicate noise levels (green: low, violet: high).}
    \label{fig:sim_observation}
\end{figure*}

We also analyzed the effects of feature track count $M$ and per-track observation number $N_i=n$ under two combined noise conditions: $1 \mathrm{px} + 1 \mathrm{ms} + 2 \mathrm{deg/s}$ and $2 \mathrm{px} + 2 \mathrm{ms} + 5 \mathrm{deg/s}$. In general, track count ($M$) correlates with spatial resolution---high-resolution cameras (\eg~regular frame-based sensors) can maintain hundreds of tracks spatially. Observation count ($N_i=n$) depends on temporal resolution: event cameras, with microsecond-level precision, can densely sample a track over short time windows, enabling high $n$ values (\eg~50 observations) even in constrained durations, which explains our choice of the upper bound. As shown in~\figref{fig:sim_observation} (left), increasing tracks from 3 to 30 significantly reduces velocity errors under both noise levels, but has marginal gains beyond 30 tracks. In contrast, raising observation counts per track under fixed time windows (\figref{fig:sim_observation} right) shows limited efficacy. Thus, the interplay between $M$ and $n$ reflects sensor-specific spatiotemporal trade-offs: frame-based systems excel in spatial coverage (high $M$) with temporally sparse measurements, while event cameras leverage temporal uniformity (high $n$) despite lower track counts. This highlights how sensor architecture shapes robustness under multi-source noise.

\subsubsection{Analysis of Track Length}
\begin{figure}
    \centering
    \includegraphics[width=0.7\linewidth]{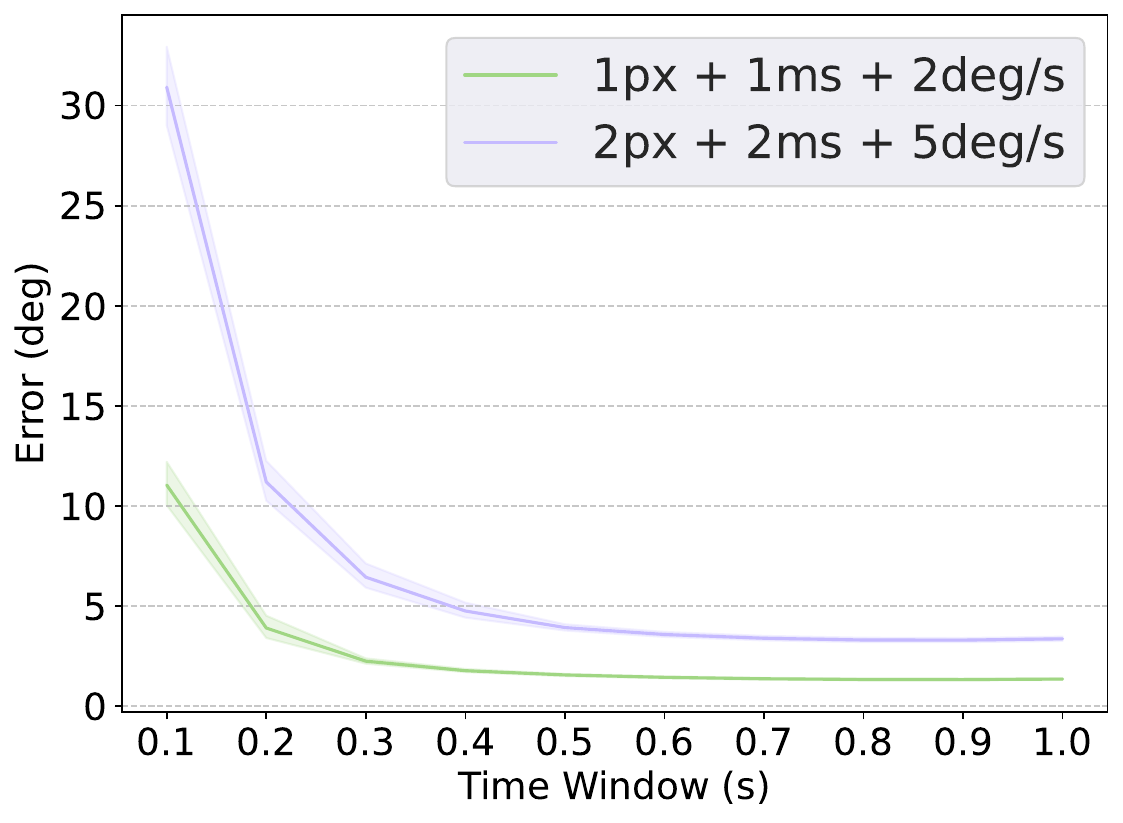}
    \vspace{-2ex}
    \caption{Velocity errors vs. time window length and noise.}
    \label{fig:sim_time_window}
\end{figure}

We further analyze the impact of the track length--determined by the temporal observation window--on velocity estimation. The time interval indicates the spatial displacement of feature tracks on the imaging plane: longer windows generally provide longer tracks, as features travel larger pixel distances under camera or scene motion. As illustrated in~\figref{fig:sim_time_window}, we indirectly control track length by varying the size of the time window. Under combined noise conditions, larger time windows consistently improve the solver's robustness. Longer tracks are able to average out high-frequency noise which enables the solver to recover stable velocity estimates. It also suggests that 3D points with smaller depth that induce larger apparent motion on the imaging plane can provide more reliable results. Notably, event cameras, which maintain temporally uniform observations, can benefit from larger time window.

\subsection{Real Experiments}
\label{sec:real-exp}
In a next step, we deploy our solver on real data collected with three different sensing modalities (as described in~\secref{sec:method}) using public datasets: The Event Camera Dataset~\cite{mueggler2017event} was collected using a DAVIS camera~\cite{brandli2014240} which provides a pixel-aligned and time-synchronized stream of global shutter ($240\times180$@24Hz) and event camera data, as well as IMU measurements. The VECtor dataset~\cite{gao2022vector} provides event camera data and global shutter frames ($1224\times1024$@30Hz) from a stereo rig, as well as IMU data. Finally, the TUM dataset~\cite{schubert2019rolling} provides rolling shutter images ($1280\times1024$@20Hz). We compare our method against eventail~\cite{gao2024eventail}, an asynchronous line-based velocity solver. It uses clusters of events, each generated by a separate line to regress velocity components, and fuses these to a full normalized velocity via velocity averaging. 
On rolling shutter images we implement a similar baseline based on eventail, termed eventail + RS. eventail + RS first extracts Canny edges~\cite{Canny86pami} from the rolling shutter images, and then treats the detected points as events with timestamps assigned according to their row index. 
We do not report acceleration estimation results on real data due to high noise sensitivity, and leave addressing this challenge for future work.

\textbf{Experimental Setup}:
For global and rolling shutter images, we employ the standard Kanade-Lucas-Tomasi Tracker~\cite{Lucas81ijcai}. For global shutter images we assign timestamps at the midpoint of the exposure time. For rolling shutter images we compute timestamps based on the row index of the feature and scanning time. For event camera data we use the recent learning-based point tracker ETAP~\cite{hamann2025motion} to generate point tracks that preserve the high temporal resolution. To ensure the numerical stability of the solver, we filter out tracks shorter than 10 pixels, as these often amplify noise. All methods operate within identical time windows to enable fair comparison. In each sequence we use angular rates measured by an IMU to rotation compensate the bearing vectors. The RANSAC pipeline is configured with a maximum iteration of 200, with each iteration sampling $M = 4$ randomly selected tracks containing $N_i = 5$ observations. A point track is classified as an inlier if its angular error falls below $5^{\circ}$. To speed up convergence, we terminate RANSAC when the inlier ratio exceeds 0.9.

\begin{table}[tbp]
\centering
\caption{Mean / median velocity error in degrees on tracks from global shutter images. * and gray: results on subset with track inlier ratio $> 0.9$. eventail + E uses events from an event camera.}
\label{tab:gs_results}
\resizebox{\linewidth}{!}{
\begin{tabular}{@{}l cc|c@{}}
\toprule
\textbf{Seq.} & eventail~\cite{gao2024eventail} + E & Ours & {\color{gray}Ours*}  \\

\midrule
\emph{desk-normal} &
22.7 / 23.4 & \textbf{15.1} / \textbf{8.5} & {\color{gray} 10.2 / 7.3}\\

\emph{sofa-normal}&
 21.9 / 17.6 & \textbf{15.9} / \textbf{7.8} & {\color{gray} 9.8 / 6.3}\\

\emph{mountain-normal}&
 25.2 / 21.4 & \textbf{17.1} / \textbf{7.5} & {\color{gray}10.9 / 6.1} \\
\midrule

\emph{shapes\_translation}
 &31.8 / 32.7 & \textbf{17.1} / \textbf{7.2}  & {\color{gray}9.9 / 6.2} \\

\emph{boxes\_translation}
 &34.8 / 34.1 & \textbf{16.5} / \textbf{11.6} & {\color{gray} 13.3 / 10.7} \\
\bottomrule
\end{tabular}}

\vspace{5pt}
\end{table}

\noindent\textbf{Application to Global Shutter Cameras}
We report the result of our method in~\tabref{tab:gs_results}. It can be seen that our method using global shutter images achieves a lower error than the eventail solver operating on events. This is due to two effects: First, the eventail solver relies on extracting events generated by 3D lines in the scene, which is limited to highly geometric structures. By contrast our method can rely on tracks which can be extracted more easily. The second effect is the use of images vs. events. Events are known to suffer from changing appearance due motion changes, leading to drift in the feature tracks~\cite{gehrig2020eklt,gehrig2018asynchronous}. We will see later that combining tracks from colocated GS cameras and events can improve results, even beyond the results using global shutter cameras alone. 
Next, we focus on the results marked with *, indicating evaluation on the subset where over 90\% of tracks are termed inliers. On this subset, errors are further reduced, showing the importance of having geometrically consistent observations.

\begin{table}[tbp]
\centering
\caption{Mean / median velocity error in degrees on tracks from rolling shutter images. * and gray: results on subset with inlier ratio $> 0.9$. eventail + RS uses canny edges from rolling shutter images. ``no correction'': no rolling shutter timestamp correction.}
\label{tab:rs_results}
\resizebox{\linewidth}{!}{
\begin{tabular}{@{}l ccc|cc@{}}
\toprule
\multirow{2}{*}{\textbf{Seq.}} & \multicolumn{3}{c|}{\textbf{with correction}} & \multicolumn{2}{c}{\textbf{no correction}}\\
\cmidrule(lr){2-6}
& eventail~\cite{gao2024eventail} + RS& Ours& {\color{gray} Ours*} & Ours & {\color{gray} Ours*}  \\
\midrule
\emph{Seq 4} & 43.8 / 40.8 & \textbf{27.5} / \textbf{20.1} & {\color{gray} 22.6 / 17.4} & 28.1 / 22.9 & {\color{gray} 22.8 / 15.7} \\
\emph{Seq 5} & 45.5 / 44.8 & \textbf{24.7} / \textbf{17.0} & {\color{gray} 19.3 / 13.8} & 27.0 / 18.4 & {\color{gray} 19.2 / 14.6} \\
\bottomrule
\end{tabular}}
\vspace{-3ex}
\end{table}

\noindent\textbf{Application to Rolling Shutter Cameras} 
We report the results of our method in~\tabref{tab:rs_results}. We see that our point solver yields a significant improvement with respect to the eventail solver. This is mainly due to the fact that eventail found few lines in the presented sequences. By contrast, our method relies on feature tracks which are more easily extracted. We also show a result without rolling shutter timestamp correction, denoted with ``no correction''. It is visible that this reduces the accuracy of the method by a few degrees, showing the benefit of correct timestamp association, and the flexibility of our method to take non-synchronized feature tracks into account. Finally, as before we see that results on the subset marked with * are better, indicating the importance of geometrically consistent tracks for estimation.

\begin{table}[tbp]
\centering
\caption{Mean / median velocity error in degrees on tracks from events. * and gray: results on subset with track inlier ratio $> 0.9$. E stands for events from an event camera. E+GS refers to our method combining tracks from events and global shutter images. Note that for - no collocated GS and event sensor are available.}
\label{tab:e_results}
\resizebox{\linewidth}{!}{
\begin{tabular}{@{}l ccccccc@{}}
\toprule
\textbf{Seq.} & eventail~\cite{gao2024eventail} + E  & Ours + E  & {\color{gray}Ours* + E} & Ours + E + GS & {\color{gray}Ours* + E + GS} \\

\midrule
\emph{desk-normal} 
 & 22.7 / 23.4 & \textbf{19.3} / \textbf{17.8} & {\color{gray} 14.2 / 14.2} & -- & {\color{gray}--} \\

\emph{sofa-normal}
 & 21.9 / 17.6 & \textbf{19.0} / \textbf{18.5} & {\color{gray}16.3 / 14.9} & -- & {\color{gray}--} \\

\emph{mountain-normal}
 & 25.2 / 21.4 & \textbf{17.1} / \textbf{16.1} & {\color{gray}16.9 / 15.8} &  -- & {\color{gray}--} \\
\midrule

\emph{shapes\_translation}
 & 31.8 / 32.7 & 16.8 / 10.1 & {\color{gray} 13.0 / 9.1} & \textbf{14.4} / \textbf{7.5} & {\color{gray}7.0 / 6.7} \\

\emph{boxes\_translation}
 & 34.3 / 34.1 & 12.6 / 10.0 & {\color{gray}12.1 / 7.7}  &  \textbf{10.3} / \textbf{8.1} & {\color{gray}9.3 / 5.9} \\
\bottomrule
\end{tabular}}

\vspace{-2ex}
\end{table}

\noindent\textbf{Application to Event-based Cameras}
Finally, we apply our method to tracks derived from an event-based camera, and show results in~\tabref{tab:e_results}.  Our point solver running on events alone outperforms eventail by $10-30\%$, and this is again due to the use of point tracks instead of lines. Interestingly, frame-based tracks yield better performance than event-based ones, particularly on the VECtor sequences. This aligns with our simulation findings where higher feature density (from frame cameras' $5\times$ resolution advantage over event cameras) improves spatial sampling. We show the benefit of combining tracks from different sensing modalities, denoted with E+GS. In particular, the sequences \emph{shapes\_translation} and \emph{boxes\_translation} were recorded with a DAVIS camera~\cite{brandli2014240} which features pixels that simultaneously record events and global shutter images. In this setting, we see that adding images significantly reduces errors. Moreover, comparing to~\tabref{tab:gs_results} we also see that adding events improves over the global shutter result, highlighting the complementarity of the sensors. This result highlights the benefit of having an asynchronous point solver that can flexibly incorporate both global shutter and event-based observations.

\section{Future Work and Conclusion}
\label{sec:conclusion}
\noindent\textbf{Future Work:} While we believe that the proposed solver makes a significant stride toward handling asynchronous tracks in an efficient way, we acknowledge its dependence on available angular rates from an IMU. Initial steps have been made in incorporating angular rate estimation into existing solvers~\cite{gao2023eventail,gao2024eventail}, but further work is needed to make these solvers efficient. Finally, we only show linear acceleration estimation in simulation (see appendix), and found significant challenges with noise on real-world data. This indicates estimation stability issues for higher-order derivatives. Future work should aim to identify and reduce the effect of noise on higher-order derivative estimation.\\
\noindent\textbf{Conclusion:} We present a linear N-point solver for recovering structure and linear motion from asynchronous feature tracks. It generalizes solvers that rely on additional structure constraints such as points lying on a line, or time constraints, such as assuming synchronized timestamps. We showed experimentally that the motions recovered by our solver are more accurate than those produced by previous work, and also more robust in natural, line-deprived environments. We believe that our solver sets the stage for many new innovations to come by enabling the seamless integration of asynchronous feature tracks into geometric solvers.  

\section*{Acknowledgments}
\label{sec:acknowledgments}

This research has been supported by project 62250610225 by the Natural Science Foundation of China, as well as projects 22DZ1201900, and dfycbj-1 by the Natural Science Foundation of Shanghai and project 225354 of the Swiss National Science Foundation.

{
  \small
  \bibliographystyle{ieeenat_fullname}
  \bibliography{main}
}

\section{Appendix}

\subsection{Explicit Matrix Formulas}
In the main text, we discuss the use of matrices $\mathbf{M}_A,\mathbf{M}_B,\mathbf{M}_C$ defined via
\begin{equation}    \underbrace{\mathbf{A}^\intercal\mathbf{A}}_{\doteq \mathbf{M}}\mathbf{x}=
    \begin{bmatrix}
        \mathbf{M}_{A} & \mathbf{M}_{B} \\
        \mathbf{M}_{B}^\intercal & \mathbf{M}_{D}
    \end{bmatrix}
    \begin{bmatrix}
        \mathbf{P}_{1:M} \\
        \mathbf{v}  
    \end{bmatrix} 
    = \mathbf{0}_{(3M+3)\times 1},
\end{equation}
where $\mathbf{A}$ is defined via 
\begin{equation}
    \underbrace{
    \begin{bmatrix}
        \mathbf{F}_1 & & & & \mathbf{G}_1 \\
        & \mathbf{F}_2 & & & \mathbf{G}_2 \\
        & & \ddots & & \vdots \\
        & & & \mathbf{F}_M & \mathbf{G}_M
    \end{bmatrix}}_{\doteq \mathbf{A}\in \mathbb{R}^{3N\times (3M+3) }}
    \underbrace{\begin{bmatrix}
        \mathbf{P}_1 \\
        \mathbf{P}_2 \\
        \vdots \\
        \mathbf{P}_M \\
        \mathbf{v}
    \end{bmatrix}}_{\doteq \mathbf{x}\in\mathbb{R}^{3M+3}}
    = \mathbf{0}_{3N\times1}
\end{equation}
and each $\mathbf{F}_i,\mathbf{G}_i$ is defined as 
\begin{equation}
    \underbrace{\begin{bmatrix}
        \left[\mathbf{f}'_{ij}\right]_\times & -t'_{ij} \left[\mathbf{f}'_{ij}\right]_\times \\ 
        \vdots & \vdots \\
        \left[\mathbf{f}'_{iN_i}\right]_\times & -t'_{iN_i} \left[\mathbf{f}'_{iN_i}\right]_\times
    \end{bmatrix}}_{\doteq\begin{bmatrix}
        \mathbf{F}_i&\mathbf{G}_i
    \end{bmatrix}\in\mathbb{R}^{3N_i\times 6}} \begin{bmatrix}
        \mathbf{P}_i\\\mathbf{v}
    \end{bmatrix}=\mathbf{0}_{3N_i\times 1}
\end{equation}
We thus see that 
\begin{align}
    \mathbf{M}_A &= \begin{bmatrix}
        \mathbf{F}_1^\intercal \mathbf{F}_1&&&\\ 
        &\mathbf{F}_2^\intercal \mathbf{F}_2&&\\
        &&\ddots&\\
        &&&\mathbf{F}_M^\intercal \mathbf{F}_M
    \end{bmatrix}\\
    \mathbf{M}_B &= \begin{bmatrix}
        \mathbf{F}_1^\intercal\mathbf{G}_1\\
        \mathbf{F}_2^\intercal\mathbf{G}_2\\
        \vdots \\
        \mathbf{F}_M^\intercal\mathbf{G}_M
    \end{bmatrix}\\
    \mathbf{M}_D&=\sum_{i=1}^M \mathbf{G}^\intercal_i\mathbf{G}_i
\end{align}
where 
\begin{align}
    \mathbf{F}_i^\intercal \mathbf{F}_i &= -\sum_{j=1}^{N_i} [\mathbf{f}'_{ij}]^2_\times\\
    \mathbf{F}_i^\intercal \mathbf{G}_i &= \sum_{j=1}^{N_i}t'_{ij} [\mathbf{f}'_{ij}]^2_\times\\
    \mathbf{G}_i^\intercal \mathbf{G}_i &-= \sum_{j=1}^{N_i} {t'_{ij}}^2[\mathbf{f}'_{ij}]^2_\times
\end{align}
share significant computation due to the products $[\mathbf{f}'_{ij}]^2_\times$.
Finally, computing the Shur complement 
\begin{equation}
    \mathbf{B} = \mathbf{M}_D - \mathbf{M}_B^\intercal \mathbf{M}^{-1}_A \mathbf{M}_B
\end{equation}
we find that 
\begin{equation}
    \mathbf{B} = \sum_{i=1}^M \mathbf{G}^\intercal_i\mathbf{G}_i - \mathbf{G}_i^\intercal \mathbf{F}_i(\mathbf{F}_i^\intercal\mathbf{F}_i)^{-1}\mathbf{F}_i^\intercal \mathbf{G}_i
\end{equation}

\subsection{Rank of \texorpdfstring{$\mathbf{B}$}{TEXT}}
Assuming a full rank of $\mathbf{F}_i$, we know that each $\mathbf{G}_i$ has full rank, since $\mathbf{G}_i=\text{diag}(t'_{i1}\mathbf{I}_{3\times3},...,t'_{iN_i}\mathbf{I}_{3\times3})\mathbf{F}_i$, and the diagonal matrix has full rank for $t'_{ij}\neq 0$. Thus, assume given the SVD of $\mathbf{F}_i=\mathbf{U}_{i}\boldsymbol{\Sigma}_{i}\mathbf{V}_i^\intercal$. Due to the full rank we have that $\boldsymbol{\Sigma}^{-1}_i$ exists.

Thus, the above formula for $\mathbf{B}$ becomes 
\begin{equation}
    \mathbf{B} = \sum_{i=1}^M \mathbf{G}^\intercal_i\mathbf{G}_i - \mathbf{G}_i^\intercal \mathbf{U}_i\mathbf{U}_i^\intercal \mathbf{G}_i
\end{equation}
which can be factored into 
\begin{equation}
    \mathbf{B} = \mathbf{G}^\intercal \left(\mathbf{I}_{3N\times 3N} - \mathbf{U}\mathbf{U}^\intercal\right)\mathbf{G}
\end{equation}
where 
\begin{align}
    \mathbf{G} = \begin{bmatrix}
        \mathbf{G}_1\\\vdots\\\mathbf{G}_M  \end{bmatrix}\quad
    \mathbf{U}=\begin{bmatrix}
    \mathbf{U}_1&&\\
        &\ddots&\\
        &&\mathbf{U}_M    \end{bmatrix}
\end{align}
To check that $\mathbf{B}$ has full rank we assume a solution $\mathbf{\hat{v}}$ to $\mathbf{B}\hat{\mathbf{v}}=\mathbf{0}_{3\times 1}$, and then need to enforce that it must be zero. Then 
\begin{equation}
    \mathbf{G}^\intercal \left(\mathbf{I}_{3N\times 3N} - \mathbf{U}\mathbf{U}^\intercal\right)\mathbf{G}\hat{\mathbf{v}}=\mathbf{0}_{3\times 1}
\end{equation}
However, since $\mathbf{G}^\intercal$ has full rank, this implies 
\begin{equation}
\left(\mathbf{I}_{3N\times 3N} - \mathbf{U}\mathbf{U}^\intercal\right)\mathbf{G}\hat{\mathbf{v}}=\mathbf{0}_{3\times 1}
\end{equation}
in other words $\mathbf{G}\hat{\mathbf{v}}$ must be in the null space of  $\mathbf{I}_{3N\times 3N} - \mathbf{U}\mathbf{U}^\intercal$. This null space is exactly the column space of $\mathbf{U}$, \textit{i.e. } vectors of the form $\mathbf{U}\boldsymbol{\lambda}$ with $\boldsymbol{\lambda}\in\mathbb{R}^{3M}$. So 
\begin{equation}
    \mathbf{G}\mathbf{\hat{v}}=\mathbf{U}\boldsymbol{\lambda}
\end{equation} 
or equivalently for each block 
\begin{equation}
    \mathbf{G}_i\mathbf{\hat{v}}=\mathbf{U}_i\boldsymbol{\lambda}_i
\end{equation} 
which can be converted into the equation 
\begin{equation}
\begin{bmatrix}
    \mathbf{G}&-\mathbf{U}_i
\end{bmatrix}\begin{bmatrix}
    \mathbf{\hat{v}}\\\boldsymbol{\lambda}_i
\end{bmatrix}    =
    \mathbf{0}_{3N\times 1}
\end{equation}
If any of the matrices on the left-hand side have full rank, this implies that $\mathbf{\hat{v}}=\mathbf{0}_{3\times 1}$, and thus that $\mathbf{B}$ has full rank.

\subsection{Connection to Epipolar Constraint}
Here we show that the point incidence relation is related to the epipolar constraint used in the well-known 5-point or 8-point algorithms. This constraint is formed from point correspondences $\mathbf{x}_{ij}$ across two views at times $t_1,t_0$. This means in our setting, for each 3D point we have $N_i=2$, and $j=0,1$. Moreover, all point in one view are synchronized, \emph{i.e.} $t_{i0}=t_0$ and $t_{i1}=t_1$ for all $i=1,...,M$. We are left with two sets of equations 
\begin{align}
    \left[\mathbf{f}'_{i0}\right]_\times \mathbf{P}_{i} - t'_{0}\left[\mathbf{f}'_{i0}\right]_\times\mathbf{v} &= \mathbf{0}_{3\times 1}\\
    \left[\mathbf{f}'_{i1}\right]_\times \mathbf{P}_{i} - t'_{1}\left[\mathbf{f}'_{i1}\right]_\times\mathbf{v} &= \mathbf{0}_{3\times 1}
\end{align}
where $t'_0=t_0-t_s$ and $t'_1=t_1-t_s$. Without loss of generality we will set $t_s=t_0$ so that $t'_0=0$, $t'_1=t_1-t_0$. Introducing translation $\mathbf{t}\doteq\mathbf{v}(t_1-t_0)$ yields 
\begin{align}
    \left[\mathbf{f}'_{i0}\right]_\times \mathbf{P}_{i} &= \mathbf{0}_{3\times 1}\\
    \left[\mathbf{f}'_{i1}\right]_\times \mathbf{P}_{i} - \left[\mathbf{f}'_{i1}\right]_\times \mathbf{t} &= \mathbf{0}_{3\times 1}
\end{align}
Next we left-multiply the second set of equations by ${\mathbf{f}'}^\intercal_{i0}$ yielding 
\begin{align}
{\mathbf{f}'}^\intercal_{i0}\left[\mathbf{f}'_{i1}\right]_\times \mathbf{P}_{i} - {\mathbf{f}'}^\intercal_{i0}\left[\mathbf{f}'_{i1}\right]_\times \mathbf{t} &= 0
\end{align}
We cycle both triple products to get 
\begin{align}
\underbrace{{\mathbf{f}'}^\intercal_{i1}\left[\mathbf{P}_{i}\right]_\times{\mathbf{f}'}_{i0}}_{=0}   - {\mathbf{f}'}^\intercal_{i1}\left[\mathbf{t}\right]_\times\mathbf{f}'_{i0} &= 0
\end{align}
The first term is 0 due to $\left[\mathbf{f}'_{i0}\right]_\times \mathbf{P}_{i} = \mathbf{0}_{3\times 1}$ and we are thus left with 
\begin{align}
{\mathbf{f}'}^\intercal_{i1}\left[\mathbf{t}\right]_\times\mathbf{f}'_{i0} &= 0
\end{align}
Next we substitute $\mathbf{f}'_{i0}=\mathbf{I}_{3\times 3}\mathbf{f}_{i0}$ and $\mathbf{f}'_{i1}=\mathbf{R}(t_{1})\mathbf{f}_{i0}$. We call $\mathbf{R}\doteq \mathbf{R}(t_{1})$, then 
\begin{align}
{\mathbf{f}}^\intercal_{i1}\underbrace{\mathbf{R}^\intercal\left[\mathbf{t}\right]_\times}_{\doteq \mathbf{E}}\mathbf{f}_{i0} &= 0
\end{align}
The underlined matrix is also called the essential matrix, and the above constraint is exactly the epipolar constraint.

\subsection{Connection to Line Solver}
\label{sec:connection}
The line solver in \cite{gao2024eventail} makes the assumption that points $\mathbf{P}_i$ lie on a line positioned at unit depth. We can thus write the position of $\mathbf{P}_i$ as a linear combination of unit vectors $\mathbf{e}_1^\ell$ and $\mathbf{e}_3^\ell$ as 
\begin{equation}
    \mathbf{P}_i = p_{i,1}\mathbf{e}_1^\ell - \mathbf{e}_3^\ell.
\end{equation}
Here $\mathbf{e}_3^\ell$ points from the closest point on the line to the origin, and $\mathbf{e}_1^\ell$ points in the direction of the line. Further define the unit vector $\mathbf{e}_2^{\ell}=\mathbf{e}_3^\ell\times \mathbf{e}_1^\ell$. We start off by multiplying the original incidence relation from the left with ${\mathbf{e}_1^\ell}^\intercal $ yielding 
\begin{equation}
    {\mathbf{e}_1^\ell}^\intercal  [\mathbf{f}_{ij}]_\times\mathbf{P}_i - t'_{ij}{\mathbf{e}_1^\ell}^\intercal[\mathbf{f}_{ij}]_\times\mathbf{v}=0.
\end{equation}
Next, we use $\mathbf{a}^\intercal[\mathbf{b}]_\times \mathbf{c}=\mathbf{c}^\intercal[\mathbf{a}]_\times \mathbf{b}$ to arrive at 
\begin{equation}
    \mathbf{f}_{ij}^\intercal \mathbf{e}_2^\ell - t'_{ij}\mathbf{f}_{ij}^\intercal(\mathbf{e}_1^\ell\times \mathbf{v})=0.
\end{equation}
Finally, we express the linear velocity in the basis $\mathbf{e}_1^\ell,\mathbf{e}_2^\ell,\mathbf{e}_3^\ell$, \textit{i.e.}
\begin{equation}
    \mathbf{v}=u_x\mathbf{e}_1^\ell + u_y\mathbf{e}_2^\ell+u_z\mathbf{e}_3^\ell
\end{equation}
inserting above, and expanding the cross product we have that 
\begin{equation}
    \mathbf{f}_{ij}^\intercal \mathbf{e}_2^\ell + t'_{ij}\mathbf{f}_{ij}^\intercal(u_z\mathbf{e}_2^\ell-u_y\mathbf{e}_3^\ell)=0.
\end{equation}
This is exactly Eq. 6 in \cite{gao2024eventail}, which demonstrates that the line solver is a special case of the point solver described here.

\subsection{Arbitrary Taylor Expansions}
In what follows, we will expand the camera motion as an $S$ order Taylor Series: 
\begin{align}
    \mathbf{R}(t_{ij}) &\approx \text{exp}\left(\left[\sum_{s=1}^S\frac{\boldsymbol{\omega}^{(s)}{t'}^s_{ij}}{s!}\right]_{\times}\right)\\
    \mathbf{p}(t_{ij}) &\approx \sum_{s=1}^S \frac{\mathbf{v}^{(s)}{t'}^s_{ij}}{s!}.
\end{align}
Here we denote $\boldsymbol{\omega}^{(s)}$ the $s$ order angular rate, and $\mathbf{v}^{(s)}$ the $s$ order linear rate. Again we will focus on recovering the linear rates, and leave the angular rates as given.
Inserting these definitions into the incidence relations yields the linear system 

\begin{equation}
    \left[\mathbf{f}'_{ij}\right]_\times \mathbf{P}_{i} - \sum_{s=1}^S \frac{{t'}^s_{ij}}{s!}\left[\mathbf{f}'_{ij}\right]_\times\mathbf{v}^{(s)} = \mathbf{0}_{3\times 1}
\end{equation}
We stack all such constraints that involve the point $\mathbf{P}_i$ into a single system of equations.
\small
\begin{equation}
    \underbrace{\begin{bmatrix}
        \left[\mathbf{f}'_{i1}\right]_\times & -t'_{i1} \left[\mathbf{f}'_{i1}\right]_\times &\hdots & -\frac{{t'}^S_{i1}}{S!} \left[\mathbf{f}'_{i1}\right]_\times\\ 
        \vdots & \vdots &\ddots &\vdots\\
        \left[\mathbf{f}'_{iN_i}\right]_\times & -t'_{iN_i} \left[\mathbf{f}'_{iN_i}\right]_\times&\hdots &-\frac{{t'}^S_{iN_i}}{S!} \left[\mathbf{f}'_{iN_i}\right]_\times
    \end{bmatrix}}_{\doteq\begin{bmatrix}
        \mathbf{F}_i&\mathbf{G_i^{(1)}}&\hdots & \mathbf{G_i^{(S)}}
    \end{bmatrix}\in\mathbb{R}^{3N_i\times (3+3S)}} \begin{bmatrix}
        \mathbf{P}_i\\\mathbf{v}^{(1)}\\\vdots \\\mathbf{v}^{(S)}
    \end{bmatrix}=\mathbf{0}_{3N_i\times 1}
\end{equation}
where $\mathbf{G}^{(s)}_i\in\mathbb{R}^{3N_i\times 3}$. For $S=1$ we recover the old case. Denoting 
\begin{align}
    \mathbf{v}\doteq \begin{bmatrix}
    \mathbf{v}^{(1)}\\\vdots \\\mathbf{v}^{(S)}
    \end{bmatrix}\quad\mathbf{G}_i\doteq \begin{bmatrix}
\mathbf{G_i^{(1)}}&\hdots & \mathbf{G_i^{(S)}}
    \end{bmatrix}
\end{align}
we arrive at the same algorithm as before. However, crucially, $\mathbf{A}\in\mathbb{R}^{3N\times (3M+3S)}$ and the Shur complement reduces $\mathbf{A}^\intercal \mathbf{A}$ to a matrix $\mathbf{B}\in\mathbb{R}^{3S\times 3S}$. The solution duality and degeneracy remains the same. However, the minimality discussion needs to be adjusted since more unknowns are now introduced.

\noindent\textbf{Minimality:} The total system has $2N$ constraints, with $3M+3S-1$ unknowns (including scale ambiguity). This means that  
\begin{equation}
    N\geq \left\lceil \frac{3M+3S-1}{2} \right\rceil
\end{equation}
At the same time, stability in the solver requires 
\begin{equation}
    N\geq 2M
\end{equation}
we will try to find out when a minimal system arises, \emph{i.e.} when $\left\lceil \frac{3M+3S-1}{2} \right\rceil=2M$. We will treat two cases: \\
\noindent\textbf{Case 1:} Let $h\doteq 3M+3S-1$ be odd with $h=2k+1$ and $k=\frac{h-1}{2}$ for some $k$, then the left side of the above equation becomes 
\begin{align}
    \left\lceil \frac{h}{2} \right\rceil&=k+1\\
    &=\frac{3M+3S-2}{2}+1\\
    &=\frac{3M+3S}{2}
\end{align}
Setting this equal to $2M$ yields 
\begin{equation}
    M=3S
\end{equation}
We thus find that configurations $(S,M)=(s,3s)$ for $s=1,2,...$ yield minimal systems. Setting $N=2M=6s$ with $N_i=2$ yields a total of $12s$ equations with $12s$ unknowns coming from $3s$ 3D points and $6s$ observations.  \\ 
\noindent\textbf{Case 2:} Let $h\doteq 3M+3S-1$ be even, so that $h=2k$ and 
\begin{align}
    \left\lceil \frac{h}{2}\right\rceil&=k\\
    &=\frac{3M+3S-1}{2}.
\end{align}
Setting this equal to $2M$ we have that 
\begin{equation}
    M=3S-1.
\end{equation}
As a result, configurations  $(S,M)=(s,3s-1)$ for $s=1,2,...$ also yield minimal systems. In particular, setting $N=2M$ with $N_i=2$ we have a system of $12s-4$ equations with $12s-4$ unknowns coming from $3s-1$ 3D points, and $6s-2$ observations.

Checking the above equations for $S=1$ recovers the minimal 4 and 6 point algorithms discussed in the main text.

\subsubsection{Experiments with Acceleration Aware Solver}
\begin{figure*}
    \centering
    \includegraphics[width=\linewidth]{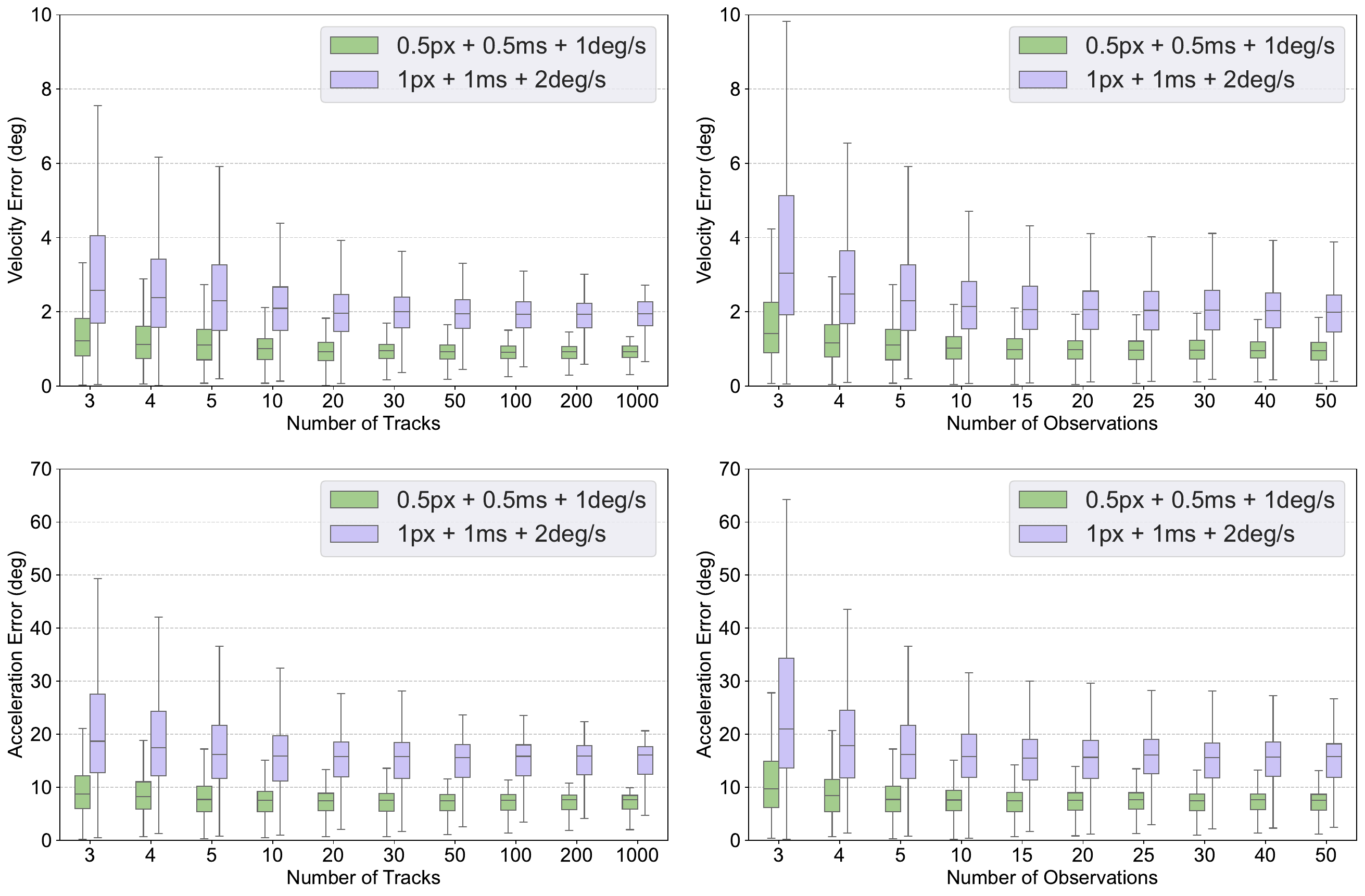}
    \caption{Analysis of acceleration-aware point solver on the number of feature tracks (left) and observations (right) under different combined noise level.}
    \label{fig:sim_observation_accel}
\end{figure*}

\begin{figure*}
    \centering
    \includegraphics[width=\linewidth]{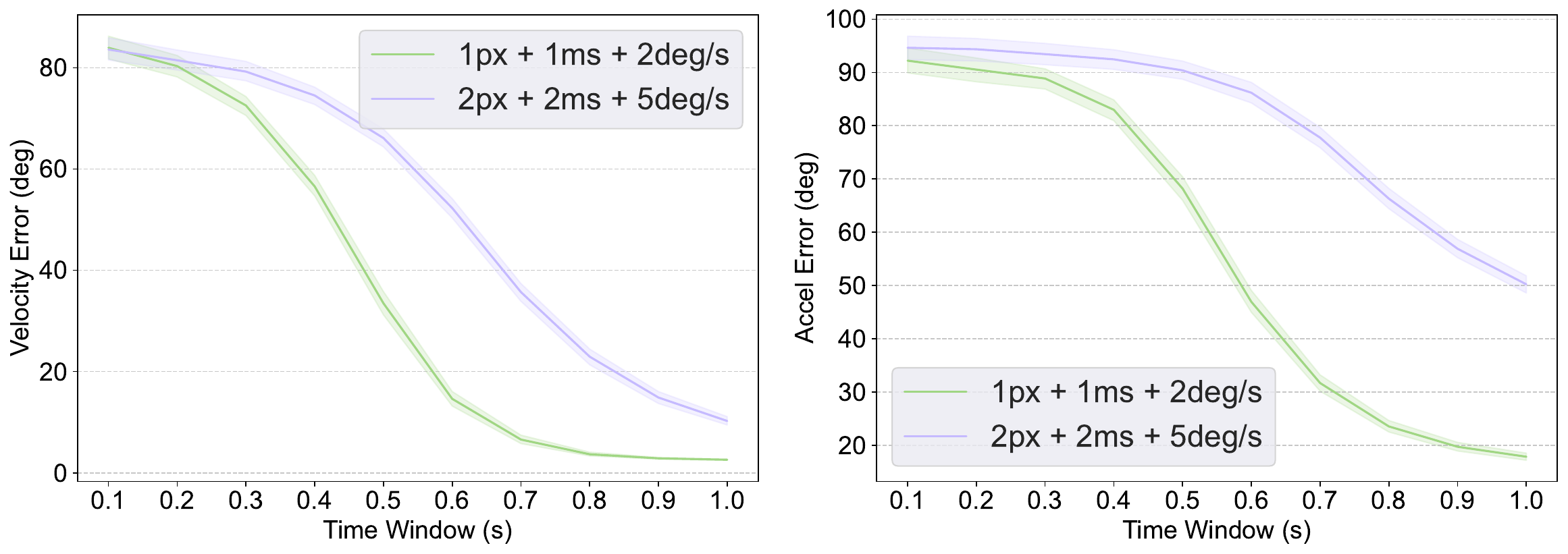}
    \caption{Analysis of the impact of time window on acceleration-aware solver under different noise level.}
    \label{fig:sim_time_window_accel}
\end{figure*}

To extend our methodology to acceleration-aware motion estimation, we conducted simulation experiments as described in Sec.4.1 with a 1-second time window. The results shown in~\figref{fig:sim_observation_accel} and \figref{fig:sim_time_window_accel} reveal that introducing acceleration parameters (adding three degrees of freedom) systematically amplifies the solver's noise sensitivity. Notably, acceleration estimation exhibits 3-4 times higher error susceptibility than velocity under identical noise conditions. This effect arises because the acceleration term's coefficient scales quadratically with the timestamp in the motion model. Nevertheless, it can be mitigated by fusing IMU acceleration data to bootstrap initial velocity estimation.

\subsection{Extending the Acceleration-aware Solver}
The above formulation for $S=2$ expands the camera motion up to acceleration yielding the incidence relation 
\begin{equation}
    \left[\mathbf{f}'_{ij}\right]_\times \mathbf{P}_{i} -  {t'}_{ij}\left[\mathbf{f}'_{ij}\right]_\times\mathbf{v}-\frac{1}{2}{t'}^2_{ij}\left[\mathbf{f}'_{ij}\right]_\times\mathbf{a} = \mathbf{0}_{3\times 1}
\end{equation}
Instead of solving for the unknown $\mathbf{a}$ we can assume it given by an IMU (which we already use to estimate angular velocity). For simplicity, we assume it to be constant over the time interval. We thus find that our system transforms into an in-homogeneous system : 
\begin{equation}
    \left[\mathbf{f}'_{ij}\right]_\times \mathbf{P}_{i} -  {t'}_{ij}\left[\mathbf{f}'_{ij}\right]_\times\mathbf{v} = \frac{1}{2}{t'}^2_{ij}\left[\mathbf{f}'_{ij}\right]_\times\mathbf{a}
\end{equation}
stacking these equations for one track yields 
\begin{equation}
    \underbrace{\begin{bmatrix}
        \left[\mathbf{f}'_{i1}\right]_\times & -t'_{i1} \left[\mathbf{f}'_{i1}\right]_\times \\ 
        \vdots & \vdots \\
        \left[\mathbf{f}'_{iN_i}\right]_\times & -t'_{iN_i} \left[\mathbf{f}'_{iN_i}\right]_\times
    \end{bmatrix}}_{\doteq\begin{bmatrix}
        \mathbf{F}_i&\mathbf{G_i}
    \end{bmatrix}\in\mathbb{R}^{3N_i\times 6}} \begin{bmatrix}
        \mathbf{P}_i\\\mathbf{v}
    \end{bmatrix}=\underbrace{\begin{bmatrix}\frac{1}{2}{t'}^2_{i1}\left[\mathbf{f}'_{i1}\right]_\times\mathbf{a}\\\vdots\\\frac{1}{2}{t'}^2_{iN_i}\left[\mathbf{f}'_{iN_i}\right]_\times\mathbf{a}\end{bmatrix}}_{\doteq \mathbf{b}_i\in\mathbb{R}^{3N_i\times 1}}
\end{equation}
and stacking these equations for each point we get 
\begin{equation}
    \underbrace{
    \begin{bmatrix}
        \mathbf{F}_1 & & & & \mathbf{G}_1 \\
        & \mathbf{F}_2 & & & \mathbf{G}_2 \\
        & & \ddots & & \vdots \\
        & & & \mathbf{F}_M & \mathbf{G}_M
    \end{bmatrix}}_{\doteq \mathbf{A}\in \mathbb{R}^{3N\times (3M+3) }}
    \underbrace{\begin{bmatrix}
        \mathbf{P}_1 \\
        \mathbf{P}_2 \\
        \vdots \\
        \mathbf{P}_M \\
        \mathbf{v}
    \end{bmatrix}}_{\doteq \mathbf{x}\in\mathbb{R}^{3M+3}}
    = \underbrace{\begin{bmatrix}
        \mathbf{b}_1\\\vdots\\\mathbf{b}_M
    \end{bmatrix}}_{\doteq\mathbf{b}\in\mathbb{R}^{3N\times 1}}
\end{equation}
multiplying from the left with $\mathbf{A}^\intercal$ we have the system 
\begin{equation}
    \label{eq:schur-complement-suppl}
    \underbrace{\mathbf{A}^\intercal\mathbf{A}}_{\doteq \mathbf{M}}\mathbf{x}=
    \begin{bmatrix}
        \mathbf{M}_{A} & \mathbf{M}_{B} \\
        \mathbf{M}_{B}^\intercal & \mathbf{M}_{D}
    \end{bmatrix}
    \begin{bmatrix}
        \mathbf{P}_{1:M} \\
        \mathbf{v}  
    \end{bmatrix} 
    = \underbrace{\mathbf{A}^\intercal \mathbf{b}}_{\doteq \mathbf{c}}=\begin{bmatrix}
        \mathbf{c}_A\\\mathbf{c}_B
    \end{bmatrix},
\end{equation}
where we now define $\mathbf{c}\in\mathbb{R}^{(3M+3)\times 1}$ with $\mathbf{c}_A\in\mathbb{R}^{3M\times 1}$ and $\mathbf{c}_B\in\mathbb{R}^{3\times 1}$. Explicit formulas for these are 
\begin{equation}
    \mathbf{c}_A = \begin{bmatrix}
        \mathbf{F}_1^\intercal \mathbf{b}_1\\\vdots\\\mathbf{F}_M^\intercal \mathbf{b}_M
    \end{bmatrix}\quad\mathbf{c}_B=\sum_{i=1}^M\mathbf{G}^\intercal_i\mathbf{b}_i
\end{equation}
where 
\begin{align}
    \mathbf{F}_i^\intercal\mathbf{b}_i&=-\frac{1}{2}\sum_{j=1}^{N_i}{t'}^2_{ij}\left[\mathbf{f}'_{iN_i}\right]^2_\times\mathbf{a}
\end{align}
and 
\begin{align}
    \mathbf{G}_i^\intercal\mathbf{b}_i&=\frac{1}{2}\sum_{j=1}^{N_i}{t'}^3_{ij}\left[\mathbf{f}'_{iN_i}\right]^2_\times\mathbf{a}
\end{align}
Applying the Schur-complement trick to this system yields the following system for $\mathbf{v}$ 
\begin{equation}
    \mathbf{B}\mathbf{v}=\underbrace{\mathbf{c}_B-\mathbf{M}_B^\intercal \mathbf{M}_A^{-1}\mathbf{c}_A}_{\doteq \mathbf{d}}
\end{equation}
where we have defined $\mathbf{B}$ previously, and we now have introduced $\mathbf{d}\in\mathbb{R}^{3\times 1}$ with the following explicit formula
\begin{equation}
    \mathbf{d}=\sum_{j=1}^M \mathbf{G}_i^\intercal\mathbf{b}_i -  \mathbf{G}_i^\intercal\mathbf{F}_i(\mathbf{F}_i^\intercal \mathbf{F}_i)^{-1}\mathbf{F}_i^\intercal\mathbf{b}_i
\end{equation}
Solving for $\mathbf{v}$ requires simply inverting the matrix, yielding 
\begin{equation}
    \hat{\mathbf{v}}=\mathbf{B}^{-1} \mathbf{d}
\end{equation}
and back substitution to get the 3D points yields 
\begin{equation}
    \hat{\mathbf{P}}_i = (\mathbf{F}_i^\intercal\mathbf{F}_i)^{-1}\mathbf{F}_i^\intercal(\mathbf{b}_i - \mathbf{G}_i\mathbf{\hat{v}})
\end{equation}
It is important to note that, having taken the acceleration into account adds two qualitative differences between our solution to the previous cases: First, absolute scale suddenly becomes observable, meaning that the recovered $\hat{\mathbf{v}},\hat{\mathbf{P}}_i$ are in meters. Second, our method no longer has multiple solutions, making the depth check unnecessary. 

\begin{figure*}
    \centering
    \includegraphics[width=\linewidth]{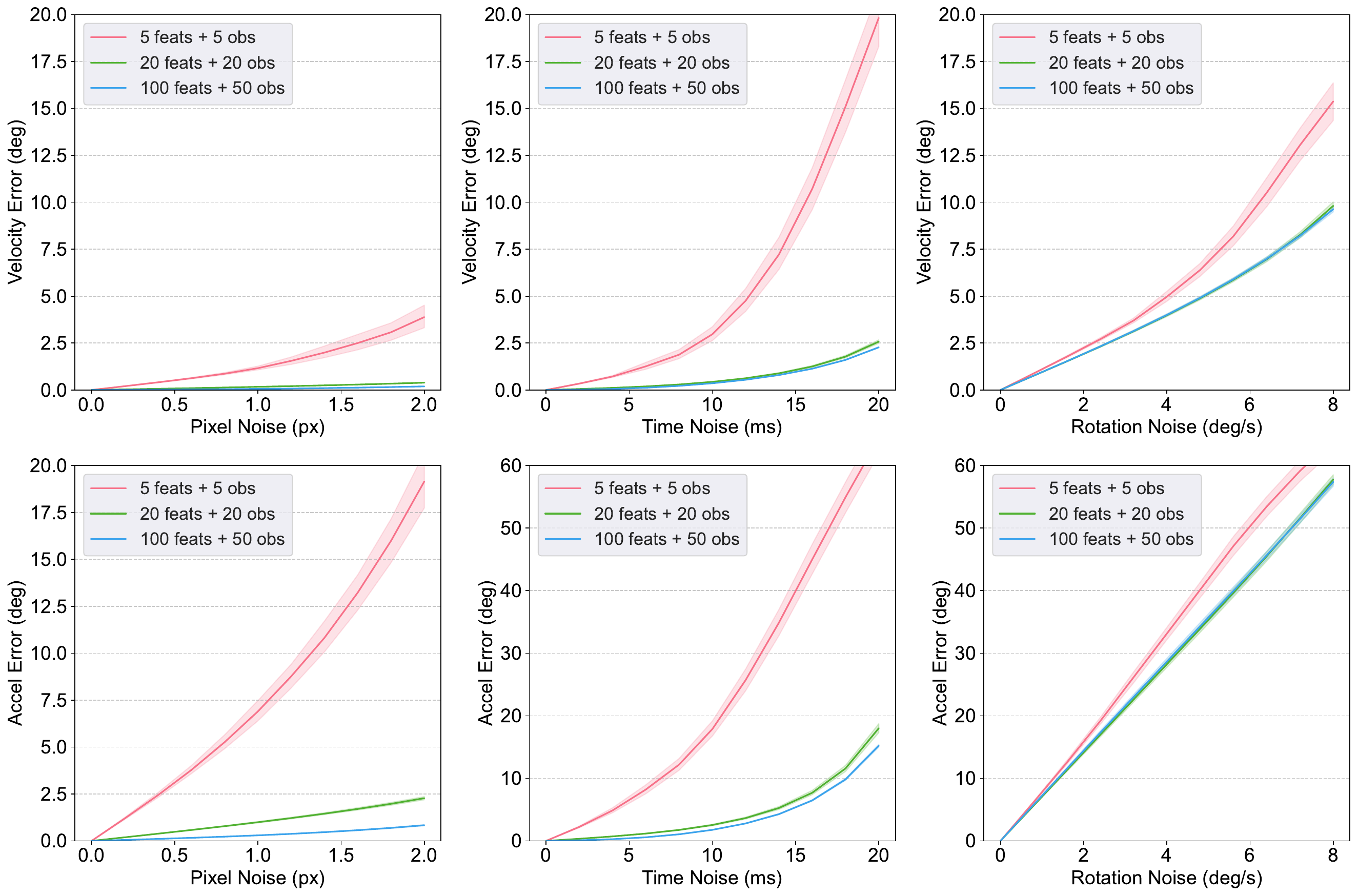}
    \caption{The robustness of acceleration-aware point solver against pixel noise (left) and timestamp jitter (middle) and rotation (right)}.
    \label{fig:sim_noise_accel}
\end{figure*}

\end{document}